\documentclass[pdflatex,sn-mathphys-num]{sn-jnl}

\usepackage[utf8]{inputenc}
\usepackage{graphicx}
\usepackage{multirow}
\usepackage{amsmath,amssymb,amsfonts}
\usepackage{amsthm}
\usepackage{mathrsfs}
\usepackage[title]{appendix}
\usepackage{xcolor}
\usepackage{textcomp}
\usepackage{manyfoot}
\usepackage{booktabs}
\usepackage{siunitx}
\usepackage{pifont}
\usepackage{url}

\theoremstyle{thmstyleone}

\theoremstyle{thmstyletwo}

\theoremstyle{thmstylethree}

\begin{document}

\title[Throwing4: Phase-Aware Pose Modeling for Throw-Distance Prediction]{Throwing4: Phase-Aware Pose Modeling for Throw-Distance Prediction in Elite Athletics}

\author[1]{\fnm{Bikash Kumar} \sur{Badatya}}\email{bikash.badatya@iitgn.ac.in}
\author[2]{\fnm{Kartike} \sur{Tiwari}}\email{2023041041@mmmut.ac.in}
\author[3]{\fnm{Jyotirmoy} \sur{Amin}}\email{jyotirmoy22@iiserbpr.ac.in}
\author*[1]{\fnm{Ravi} \sur{Hegde}}\email{hegder@iitgn.ac.in}

\affil*[1]{\orgdiv{Department of Electrical Engineering}, \orgname{Indian Institute of Technology Gandhinagar}, \orgaddress{\city{Gandhinagar}, \country{India}}}
\affil[2]{\orgdiv{Department of Electronics and Communication Engineering}, \orgname{Madan Mohan Malaviya University of Technology}, \orgaddress{\city{Gorakhpur}, \country{India}}}
\affil[3]{\orgdiv{Department of Mathematical Sciences}, \orgname{IISER Berhampur}, \orgaddress{\city{Berhampur}, \country{India}}}

\abstract{Elite athletics throwing provides a structured biomechanical action sequence together with an objective continuous outcome, namely throw distance, making it a suitable setting for performance-oriented sports analytics. In this work, we investigate whether explicit phase alignment of pose dynamics improves throw-distance prediction under realistic broadcast conditions. To support this study, we curate Throwing4, a dataset of 500 elite throwing clips spanning javelin, discus, hammer, and shot put, with corresponding two-dimensional pose sequences, frame-level phase annotations, and official distance labels. We formulate throw-distance prediction from broadcast-derived pose as a regression problem and compare full-sequence modeling against a phase-conditioned approach that partitions each attempt into four technique stages and fuses phase-specific motion representations. Experiments across multiple graph-based pose encoders show a consistent reduction in prediction error when phase structure is incorporated, indicating that phase-aligned modeling provides a more stable representation than treating the entire motion as an unstructured sequence. These findings suggest that explicit biomechanical phase information is a useful inductive prior for pose-based performance modeling in elite athletics throwing. Beyond score prediction, the dataset and protocol provide a basis for future work on fine-grained technique analysis, interpretable sports analytics, and robust modeling from broadcast video.}


\maketitle

\section{Introduction}
\label{sec:intro}


Computer vision and artificial intelligence (AI) are increasingly used in sports analysis to convert raw video into measurable information about technique, timing, and performance. Modern pipelines can track athletes, estimate body keypoints, detect critical events, localize actions in time, and summarize movement patterns from realistic broadcast footage \cite{wu2022survey,xu2025action}. This capability is important in sport because practitioners rarely need only a coarse action label; they also need to know when decisive moments occur and how the body moves through those moments. Large-scale datasets have been central to this progress. General-purpose video datasets such as UCF101 and Kinetics support broad representation learning \cite{soomro2012ucf101,kay2017kinetics}, while sports-oriented resources increasingly emphasize fine temporal structure, hierarchical action organization, and performance-sensitive supervision \cite{shao2020finegym,xu2022finediving,zhang2023logo}. At the same time, sports video remains challenging because realistic broadcasts contain camera motion, zoom, changing viewpoints, motion blur, occlusion, clutter, and editing artifacts, all of which complicate temporally precise reasoning \cite{wu2022survey,xu2025action}. This creates a persistent gap between the structured movement descriptions that coaches and analysts care about and the supervision many existing datasets or models directly provide.


A second challenge concerns how movement is measured. High-fidelity biomechanics often relies on specialized instrumentation such as motion-capture systems, markers, force plates, or wearable sensors, which can provide accurate signals but are difficult to deploy at scale in real competitions. Markerless pose estimation from ordinary red--green--blue (RGB) video offers a practical alternative because it produces compact and interpretable joint trajectories without requiring controlled laboratory setups or athlete instrumentation \cite{cao2017openpose,mmpose2020}. Once represented as skeletons, movement naturally forms a spatio-temporal graph in which body joints evolve over time, making graph-based models a natural choice for analysis \cite{yan2018stgcn,shi2019twostream,liu2020msg3d,chen2021ctrgcn}. This representation is well suited to sports in which technique is understood as an ordered sequence of phases rather than as a single indivisible motion. In such cases, explicitly incorporating phase structure may improve interpretability and may reduce the influence of variable-duration content that is less informative for the final outcome \cite{niebles2010modeling,jhuang2013towards}.

These issues are evident in athletic throwing. Throwing actions follow an ordered biomechanical progression, while performance is summarized by an objective continuous outcome: the final throw distance. Elite competitions are a suitable setting for research because technique is more standardized, attempts are governed by clear rules, and outcome labels are officially recorded. However, despite recent progress in sports datasets and pose-based modeling, there is still no widely adopted resource that jointly aligns elite throwing broadcasts with identity-consistent pose tracks, frame-level phase boundaries, and official distance outcomes across multiple throwing disciplines. This motivates the present study. We formulate elite throwing score prediction from broadcast-derived pose as a phase-aware performance modeling problem, curate a multi-discipline dataset to support this study, and investigate whether phase-conditioned modeling provides a measurable advantage over full-sequence regression under realistic broadcast conditions.

The main contributions of this work are as follows:
\begin{itemize}
    \item We formulate elite throwing score prediction from broadcast-derived pose as a phase-aware performance modeling problem and curate Throwing4 to support evaluation across four athletics disciplines.
    \item We introduce a reproducible pose-based prediction protocol using identity-consistent two-dimensional (2D) pose trajectories in the 16-joint Max Planck Institute for Informatics layout (MPII-16), lightweight pose normalization, and evaluation in both normalized units and metres.
    \item We study a phase-conditioned regression formulation that encodes each technique stage separately before fusion and compare it against full-sequence modeling across multiple graph-based pose encoders.
    \item We show that phase-conditioned modeling consistently reduces prediction error under the present experimental protocol, suggesting that biomechanical phase structure is a useful inductive prior for pose-based performance modeling in elite athletics throwing.
\end{itemize}

\section{Related Work}

\subsection{Sports video datasets and benchmarks}

Progress in sports vision has been strongly shaped by datasets that combine realistic viewing conditions with structured supervision. Generic action datasets such as UCF101 and Kinetics established strong foundations for video representation learning~\cite{soomro2012ucf101,kay2017kinetics}, while sports-specific resources pushed the field toward domain-aware temporal reasoning. FineGym emphasizes hierarchical temporal structure~\cite{shao2020finegym}, SoccerNet supports large-scale event spotting in broadcast soccer~\cite{giancola2018soccernet,deliege2021soccernetv2}, and MultiSports addresses multi-person spatio-temporal action localization~\cite{li2021multisports}. More recent action-quality datasets, such as FineDiving and LOGO, connect video understanding with procedure-aware or score-related assessment~\cite{xu2022finediving,zhang2023logo}. These resources show a shift from coarse action classification toward temporally structured and performance-sensitive analysis. However, relatively few datasets jointly provide elite athletics throwing videos, pose trajectories, phase boundaries, and official continuous distance labels.

\subsection{Fine-grained temporal structure and phase supervision}

Many sports actions are better understood as ordered technique stages than as single holistic actions. The decomposable-motion perspective~\cite{niebles2010modeling,jhuang2013towards} aligns with coaching practice, where diagnostically important cues occur in specific temporal segments. In athletics throwing, this staged structure is explicit: preparation, force generation, release, and recovery are not interchangeable parts of the motion. Prior work on javelin phase segmentation has shown that biomechanical phase boundaries can be detected from motion dynamics and can support temporally localized analysis~\cite{badatya2025biomechanical}. Related unsupervised and weakly supervised segmentation methods also indicate that temporal embeddings, clustering, and transport-based alignment can recover structured action progression without dense labels~\cite{kukleva2019unsupervised,kumar2022unsupervised,xu2024temporally}. These studies motivate the use of phase structure as both an interpretability mechanism and an inductive bias for performance modeling.

\subsection{Pose-based temporal modeling and biomechanical performance prediction}

Markerless pose provides a compact and interpretable representation for sports movement because it captures joint trajectories from ordinary video without requiring laboratory instrumentation. Pose-estimation toolkits such as MMPose make large-scale extraction feasible~\cite{mmpose2020}, and graph-based skeleton models naturally represent the body as a spatio-temporal joint graph~\cite{yan2018stgcn,shi2019twostream,liu2020msg3d,chen2021ctrgcn}. For throwing, this representation is also biomechanically meaningful. Track-and-field throwing performance is known to depend on coordinated segmental motion, release velocity, release angle, release height, and technique consistency~\cite{zaras2021biological}. In javelin, recent kinematic studies relate release velocity to run-up and body-segment contributions~\cite{makino2023kinematic,makino2022kinematic}. In discus, release parameters are athlete-specific and strongly coupled to performance~\cite{leigh2010individualized}. Hammer-throw studies emphasize the role of release velocity and rotational dynamics~\cite{castaldi2022hammer}, while shot-put studies show that kinematic repeatability and technique-dependent variables influence throwing distance~\cite{mastalerz2022shotput}. These biomechanical findings support the central premise of this work: phase-aligned pose dynamics may provide a useful representation for predicting continuous throwing performance from broadcast-derived skeletons.

\section{Methods}
\label{sec:methods}

\subsection{Dataset curation and annotation}
\label{subsec:dataset}

Throwing4 is designed as a resource for phase-aware performance modeling in elite athletics throwing under realistic broadcast conditions. It contains curated RGB clips of complete competition
attempts, identity-consistent MPII-16 2D pose sequences extracted from each clip, frame-level phase boundaries, and the official throw distance associated with each attempt. The dataset spans
four throwing disciplines---javelin, discus, hammer, and shot put---and is compiled from major international events, including the Tokyo 2020 and Paris 2024 Olympic Games, the World Athletics
Championships (WAC) in Beijing 2015 and Oregon 2022, and the European Athletics Championships (EAC) 2024.

We identify candidate attempts by scanning long-form broadcasts and then manually review each candidate to ensure that it corresponds to a complete, valid, non-foul throw with sufficient visual
observability for pose extraction and temporal annotation. We exclude foul attempts, replay-only segments, incomplete attempts, hard scene cuts around decisive motion segments, and clips with
prolonged occlusion or incomplete athlete visibility during key moments.

For each accepted attempt, the clip is trimmed so that it begins with the earliest preparatory movement initiating the attempt and ends after release and immediate stabilization or recovery,
thereby reducing label noise and ensuring that the recorded official distance corresponds to a visually observable full attempt. The curated clips have a mean duration of 5.8~s (range 1.8--11.8~s), with per-discipline means of 5.9~s for javelin, 5.9~s for discus, 7.0~s for hammer, and 4.3~s for shot put.

\begin{figure}[t]
\centering
\includegraphics[width=\linewidth]{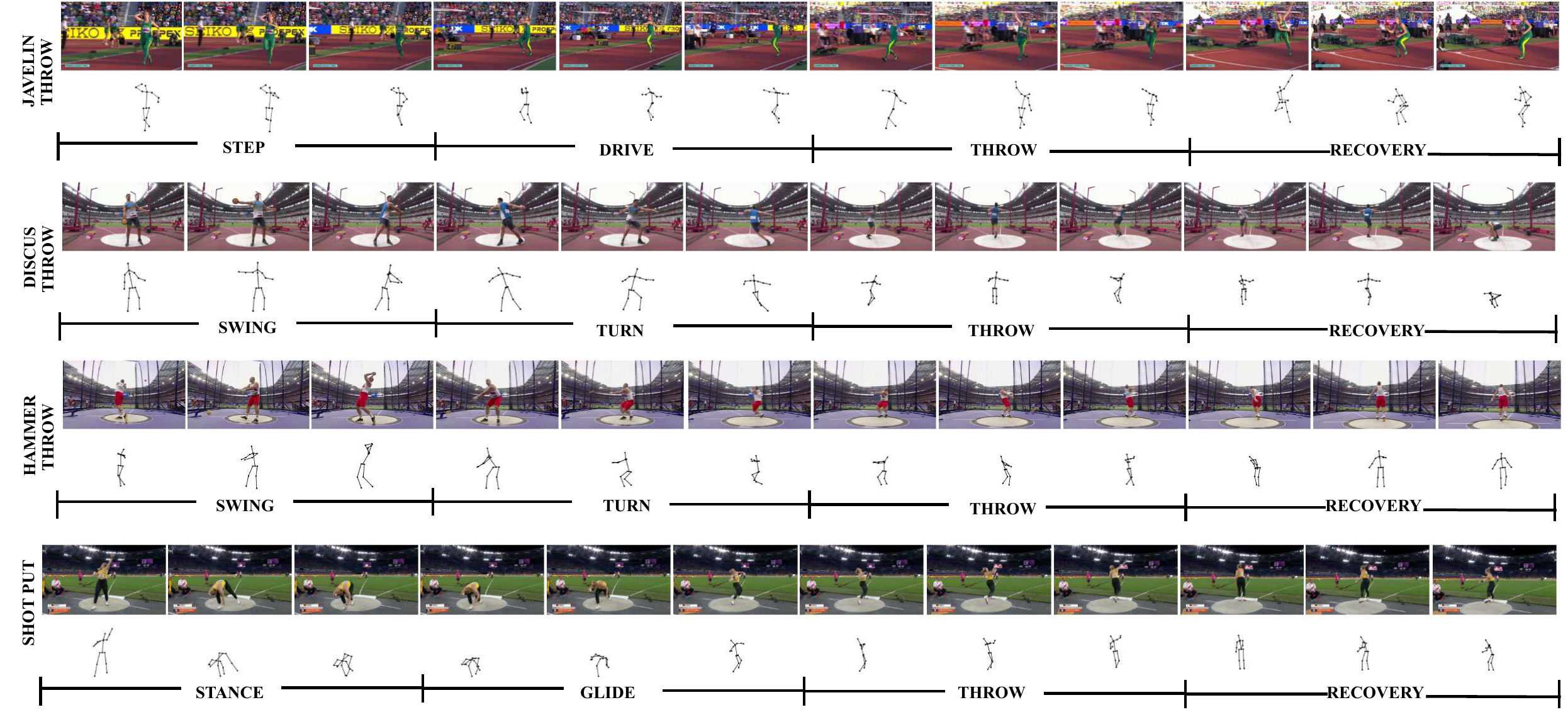}
\caption{Qualitative illustration of the four-phase technique progression for representative clips from javelin, discus, hammer, and shot put.}
\label{fig:throwing_phases_examples}
\end{figure}

Table~\ref{tab:throwing4_overview_compact} summarizes the number of curated clips per sport, the male/female split, and the source events contributing to each discipline, while
Fig.~\ref{fig:throwing_phases_examples} illustrates the four-phase technique progression used for temporal annotation. The resulting dataset contains 500 elite throwing clips in total and is
intentionally focused on valid competition attempts with reliable outcome labels rather than exhaustive coverage of all available footage.

\sisetup{detect-weight=true,detect-inline-weight=math}
\begin{table*}[t]
\centering
\caption{Throwing4 overview: number of clips per sport with male/female split (Videos (M/F)), source broadcast events, and the canonical four-phase taxonomy used for temporal annotation.}
\label{tab:throwing4_overview_compact}
\setlength{\tabcolsep}{2pt}
\renewcommand{\arraystretch}{1.1}
\tiny
\begin{tabular}{l
                l
                S[table-format=2.0]
                S[table-format=2.0]
                S[table-format=2.0]
                S[table-format=2.0]
                S[table-format=2.0]
                l l l l}
\toprule
\multirow{2}{*}{\textbf{Sport}} &
\multirow{2}{*}{\textbf{Videos (M/F)}} &
\multicolumn{1}{c}{\textbf{Tokyo}} &
\multicolumn{1}{c}{\textbf{Paris}} &
\multicolumn{2}{c}{\textbf{WAC}} &
\multicolumn{1}{c}{\textbf{EAC}} &
\multicolumn{4}{c}{\textbf{Phase}} \\
\cmidrule(lr){3-3}\cmidrule(lr){4-4}\cmidrule(lr){5-6}\cmidrule(lr){7-7}\cmidrule(lr){8-11}
& &
\textbf{2020} &
\textbf{2024} &
\textbf{Beijing 2015} &
\textbf{Oregon 2022} &
\textbf{2024} &
\textbf{1} & \textbf{2} & \textbf{3} & \textbf{4} \\
\midrule
Javelin Throw & 211 (111/100) & 45 & 82 & 27 & 16 & 41 & step   & drive & throw & recovery \\
Discus Throw  & 112 (69/43)   & 60 & 52 & {--} & {--} & {--} & swing  & turn  & throw & recovery \\
Hammer Throw  & 90  (30/60)   & 26 & 64 & {--} & {--} & {--} & swing  & turn  & throw & recovery \\
Shot Put      & 87  (55/32)   & 32 & 43 & {--} & {--} & 12 & stance & glide & throw & recovery \\
\bottomrule
\end{tabular}
\end{table*}

For every retained clip in the dataset, we manually annotate the complete video at frame resolution to obtain phase boundaries according to a compact four-phase decomposition that reflects the
canonical technique progression within each discipline. Thus, all 500 retained videos are accompanied by frame-level phase annotations, official distance labels, and corresponding
identity-consistent 2D pose sequences. These sport-specific phase names are listed in the last columns of Table~\ref{tab:throwing4_overview_compact} and visualized in
Fig.~\ref{fig:throwing_phases_examples}. The transition frames are marked by manual review at frame resolution.

These annotations serve two complementary purposes in the present study: they enable explicit phase-conditioned feature formation for regression, and they support diagnostic analysis of where
performance-relevant signal appears to concentrate within an attempt. In addition to phase boundaries, we record the official throw distance in meters for each clip as the performance label,
obtained from broadcast overlays or result panels when available. Together, the synchronized RGB clips, phase annotations, and official outcome labels provide a structured basis for studying
performance-oriented modeling in elite throwing.

Additional dataset documentation, including the folder structure, JavaScript Object Notation (JSON) annotation format, skeleton-data format, annotation procedure, athlete identifiers, attempt numbering, unique-player counts, and split-related limitations, is provided in Online Resource~1.

\subsection{Pose extraction and identity-consistent thrower tracking}
\label{subsec:pose_tracking}

\begin{figure}[t]
\centering
\includegraphics[width=\linewidth]{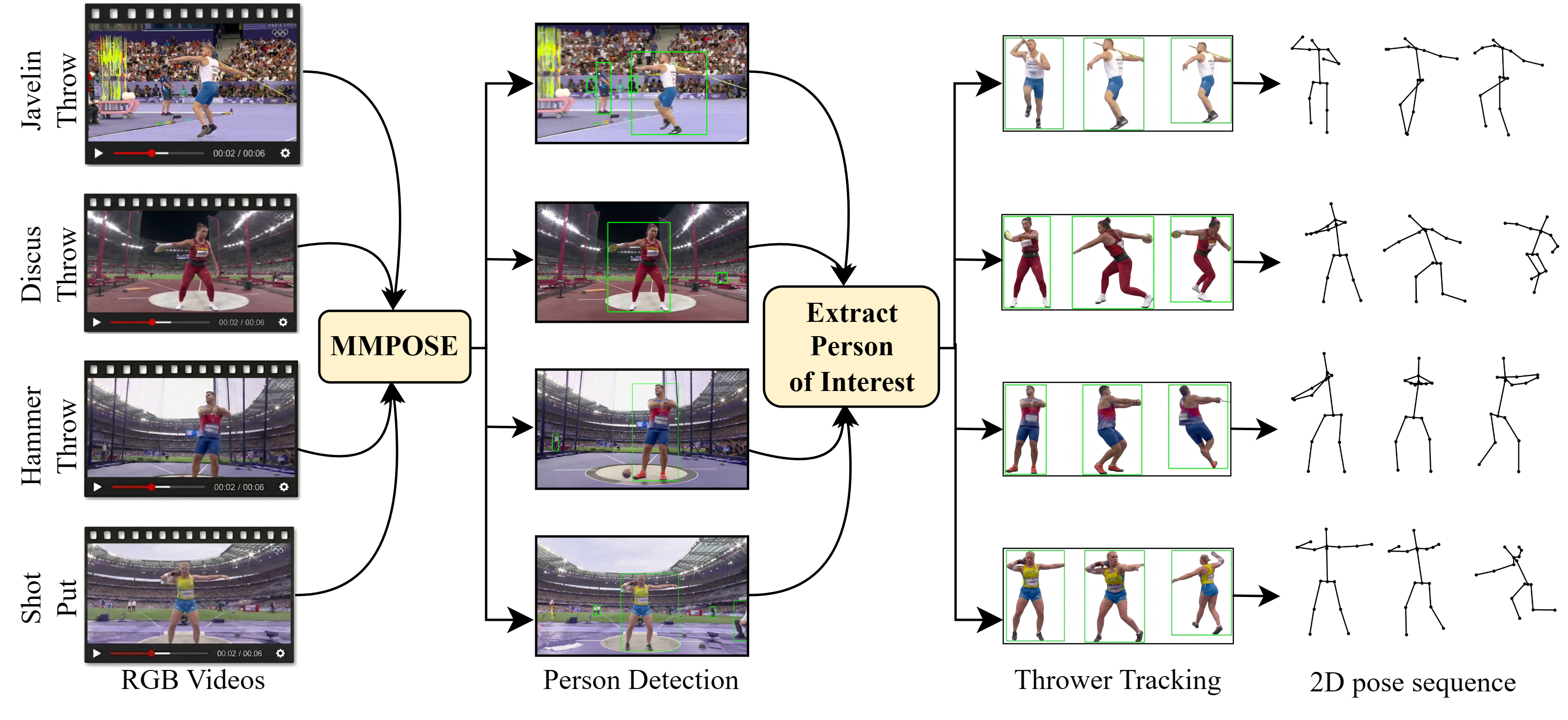}
\caption{Pose extraction pipeline in Throwing4. Broadcast RGB clips are processed with MMPose for multi-person detection, followed by identity-consistent thrower tracking to obtain an MPII-16 2D pose sequence for each clip.}
\label{fig:pose_tracking}
\end{figure}

To obtain scalable motion measurements from broadcast RGB, we extract a single identity-consistent skeleton sequence for the primary throwing athlete in each clip using a top-down pose
estimation and tracking pipeline, illustrated in Fig.~\ref{fig:pose_tracking}. Videos from all four disciplines are first processed with MMPose \cite{mmpose2020} for multi-person detection and
2D keypoint estimation. For each frame $I_t$, the detector produces a set of candidate person bounding boxes $\mathcal{B}_t$ after confidence thresholding and non-maximum suppression.

In realistic broadcast scenes, however, the frame often contains multiple people such as officials, camera operators, and nearby athletes, so naive frame-wise estimation over all detections can
introduce identity switches and temporally inconsistent pose trajectories. To address this, we construct a single tracked bounding-box sequence $\widehat{B}_t$ for the thrower of interest
using deterministic detection association.

The track is initialized from the dominant athlete box, and subsequent detections are selected by maximizing spatial overlap with the previous target subject to a gating threshold that reduces
drift under crowding, scale variation, and brief occlusion. Conditioned on $\widehat{B}_t$, MMPose produces MPII-format 2D skeletons with $J=16$ joints per frame, yielding one temporally
consistent MPII-16 pose sequence for each clip. This step converts a variable and cluttered multi-person broadcast scene into a compact athlete-centered motion representation suitable for
temporal modeling and subsequent phase-aware regression.

\subsection{Phase-aware pose modeling for throw-distance prediction}
\label{subsec:prediction}

Each attempt is represented as an MPII-16 2D pose sequence $\mathbf{P}=\{\mathbf{p}_{t,j}\}_{t=1..T,\;j=1..J}$ with $J=16$ joints and $\mathbf{p}_{t,j}\in\mathbb{R}^{2}$. To reduce nuisance variation caused by broadcast capture, including zoom, scale change, and athlete-to-camera distance, we apply a lightweight normalization of translation and scale before prediction. Let $j_p$ denote the pelvis joint index and let $t_0$ denote the first valid frame. We first remove global translation by anchoring the sequence to the pelvis at $t_0$:
\begin{equation}
\tilde{\mathbf{p}}_{t,j}=\mathbf{p}_{t,j}-\mathbf{p}_{t_0,j_p}.
\label{eq:pose_centering}
\end{equation}
For each frame, define the pelvis center $\mathbf{c}_t=\tilde{\mathbf{p}}_{t,j_p}$ and the joint offsets $\mathbf{d}_{t,j}=\tilde{\mathbf{p}}_{t,j}-\mathbf{c}_t$. The per-frame scale is then given by
\begin{equation}
r_t=\max_j \|\mathbf{d}_{t,j}\|_2,
\label{eq:pose_scale}
\end{equation}
and the normalized pose is obtained by rescaling these offsets to unit radius:
\begin{equation}
\bar{\mathbf{p}}_{t,j}=\mathbf{c}_t+\frac{\mathbf{d}_{t,j}}{r_t+\varepsilon},
\qquad \varepsilon>0.
\label{eq:pose_norm}
\end{equation}
For regression stability across disciplines with different distance ranges, the official throw distance $s$ is normalized by min--max scaling within each discipline:
\begin{equation}
y=\frac{s-s_{\min}}{s_{\max}-s_{\min}},
\qquad y\in[0,1].
\label{eq:score_minmax}
\end{equation}

\begin{figure}[t]
\centering
\includegraphics[width=\linewidth]{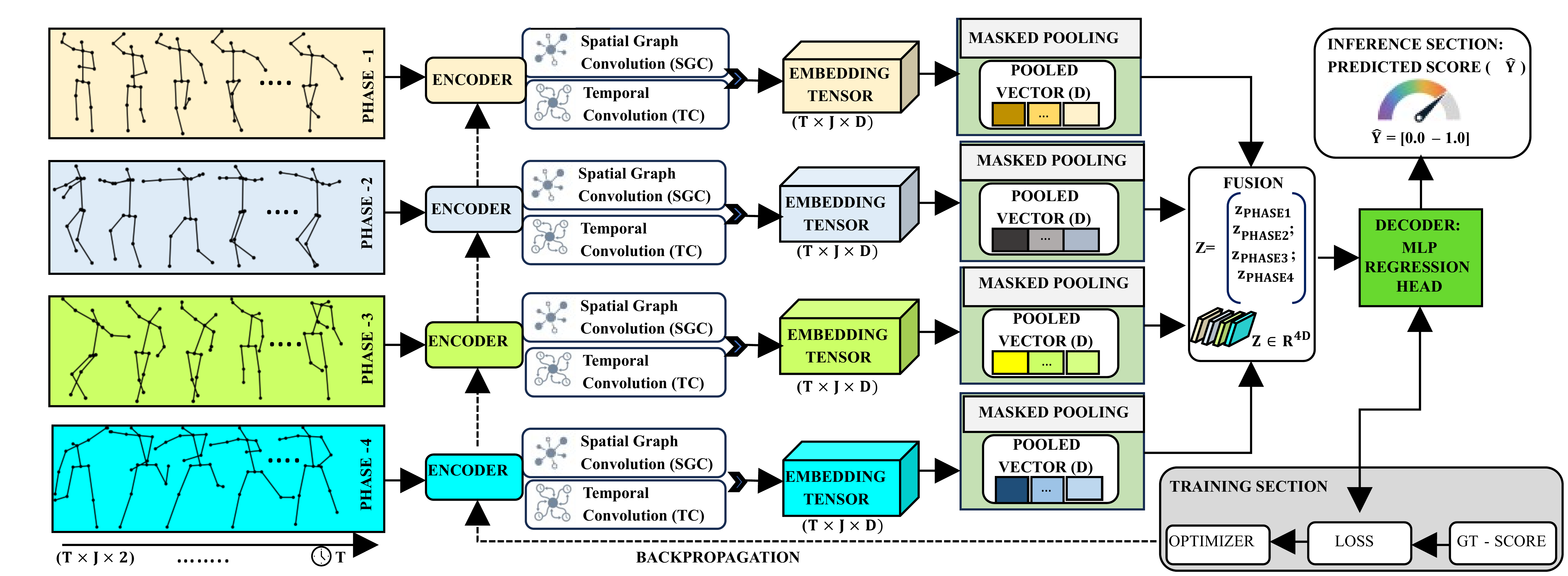}
\caption{Phase-conditioned throw-distance prediction pipeline. The pose sequence is divided into four phases, each phase is encoded by a spatio-temporal graph network, pooled with masking, and fused for regression of the normalized throw distance.}
\label{fig:score_pred_pipeline}
\end{figure}

We compare two regression formulations under a common protocol. In the first, the complete normalized pose sequence $\bar{\mathbf{P}}\in\mathbb{R}^{T\times J\times 2}$ is used directly for full-sequence modeling. Because clips have variable duration, sequences are padded to a dataset maximum length $T_{\max}$ together with a binary temporal mask $\mathbf{m}\in\{0,1\}^{T_{\max}}$. A spatio-temporal pose-graph encoder $f_{\theta}(\cdot)$ maps the padded sequence to frame-wise embeddings $\mathbf{e}_t\in\mathbb{R}^{D}$, which are aggregated by masked temporal mean pooling,
\begin{equation}
\mathbf{h}_{\mathrm{full}}
=
\frac{\sum_{t=1}^{T_{\max}} m_t\,\mathbf{e}_t}{\sum_{t=1}^{T_{\max}} m_t},
\label{eq:masked_pool_full}
\end{equation}
and a lightweight multilayer perceptron (MLP) regressor $g_{\phi}(\cdot)$ predicts the normalized score as
\begin{equation}
\hat{y}=\sigma\!\big(g_{\phi}(\mathbf{h}_{\mathrm{full}})\big)\in[0,1],
\label{eq:full_pred}
\end{equation}
where $\sigma(\cdot)$ denotes a sigmoid nonlinearity. In the second formulation, we incorporate explicit phase structure by partitioning each clip into four phase subsequences $\{\bar{P}_k\}_{k=1}^{4}$ using the manually annotated ground-truth phase boundaries. Thus, the phase-conditioned results correspond to an oracle-boundary setting in which phase boundaries are available during both training and testing. Each phase is padded to a phase-specific maximum length $L_k$ and assigned a corresponding binary mask $m_k \in \{0,1\}^{L_k}$. The same encoder is applied independently to each phase to produce embeddings $\mathbf{e}_{k,t}\in\mathbb{R}^{D}$, which are summarized by
\begin{equation}
\mathbf{h}_{k}
=
\frac{\sum_{t=1}^{L_k} m_{k,t}\,\mathbf{e}_{k,t}}{\sum_{t=1}^{L_k} m_{k,t}},
\qquad k\in\{1,2,3,4\}.
\label{eq:masked_pool_phase}
\end{equation}
The phase descriptors are concatenated as $\mathbf{h}_{\mathrm{phase}}=[\mathbf{h}_1;\mathbf{h}_2;\mathbf{h}_3;\mathbf{h}_4]\in\mathbb{R}^{4D}$ and used to predict the score through
\begin{equation}
\hat{y}=\sigma\!\big(g_{\phi}(\mathbf{h}_{\mathrm{phase}})\big)\in[0,1].
\label{eq:phase_pred}
\end{equation}
This phase-conditioned formulation encourages the model to summarize technique-aligned motion separately before fusion rather than implicitly averaging over variable-duration content across the entire clip. The architecture is illustrated in Fig.~\ref{fig:score_pred_pipeline}. For backbone attribution, Spatio-Temporal Graph Convolutional Network (STGCN) and Adaptive Graph Convolutional Network (AGCN) follow canonical skeleton-action formulations \cite{yan2018stgcn,shi2019twostream},  Attention-based Spatio-Temporal Graph Convolutional Network (ASTGCN) is adapted from the attention-based spatial-temporal graph convolutional network of Guo \emph{et al.} \cite{guo2019attention}, Graph Convolutional Network (GCN) follows the graph-convolution baseline of Kipf and Welling \cite{kipf2016semi}, and Temporal-Spatial Attention Graph Convolutional
Network (TSAGCN) follows the weakly supervised generative attention model of Shi \emph{et al.} \cite{shi2020weakly}.

\subsection{Evaluation protocol}
\label{subsec:evaluation}

Let $s_i\in\mathbb{R}$ denote the ground-truth throw distance for sample $i$ and let $\hat{y}_i$ denote the corresponding model prediction in normalized space for $i=1,\dots,N$. Using the normalization in Eq.~\eqref{eq:score_minmax}, we define
\begin{equation}
y_i=\frac{s_i-s_{\min}}{s_{\max}-s_{\min}},
\label{eq:minmax_pair}
\end{equation}
and report the mean absolute error (MAE) in normalized space:
\begin{equation}
\mathrm{MAE}=\frac{1}{N}\sum_{i=1}^{N}\left|y_i-\hat{y}_i\right|.
\label{eq:mae_norm}
\end{equation}
For interpretability in meters, we de-normalize the MAE by the score range:
\begin{equation}
\mathrm{MAE}_{\mathrm{m}}=(s_{\max}-s_{\min})\,\mathrm{MAE}.
\label{eq:mae_meter}
\end{equation}

The clips in each discipline are partitioned at the clip level into 80\% for training and the remaining 20\% for held-out testing. This partition is an attempt-level within-dataset split and does not enforce athlete-independent, competition-held-out, or event-held-out separation; repeated attempts from the same athlete may therefore appear in both partitions. The regressor receives only normalized two-dimensional skeleton sequences and does not use athlete names, RGB appearance, venue background, broadcast graphics, or performance history. All throw-distance prediction experiments follow this normalization, partitioning, and evaluation protocol.

All models were implemented in PyTorch and trained for 100 epochs with a batch size of 32. The normalized throw-distance regression objective was optimized using mean-squared error loss and the
AdamW optimizer with learning rate $1\times10^{-4}$ and weight decay $1\times10^{-4}$. No learning-rate scheduler was used. For each random split, the held-out test set was used only for final
reporting and was not used during training. Gradients were clipped to a maximum norm of 5.0. The pose-graph encoder produces a $D=64$-dimensional embedding per time step, which is pooled
either across the full sequence or within each phase before regression. Experiments were run on a workstation with an Intel Core i9-7920X CPU, 125 GB RAM, and NVIDIA GPUs including a GeForce
GTX 1080 Ti with 11 GB memory and two GeForce RTX 3050 GPUs with 8 GB memory.

The aim of the empirical study is not to identify a universally best graph architecture for all throwing events, but to evaluate whether explicit phase structure consistently improves
prediction under a common protocol across multiple model families. A separate repeated-seed AGCN analysis over five random seeds for each discipline is provided in Online Resource 1 to assess robustness beyond the primary benchmark.

\section{Results}
\label{sec:results}

\begin{figure}[t]
\centering
\includegraphics[width=\linewidth]{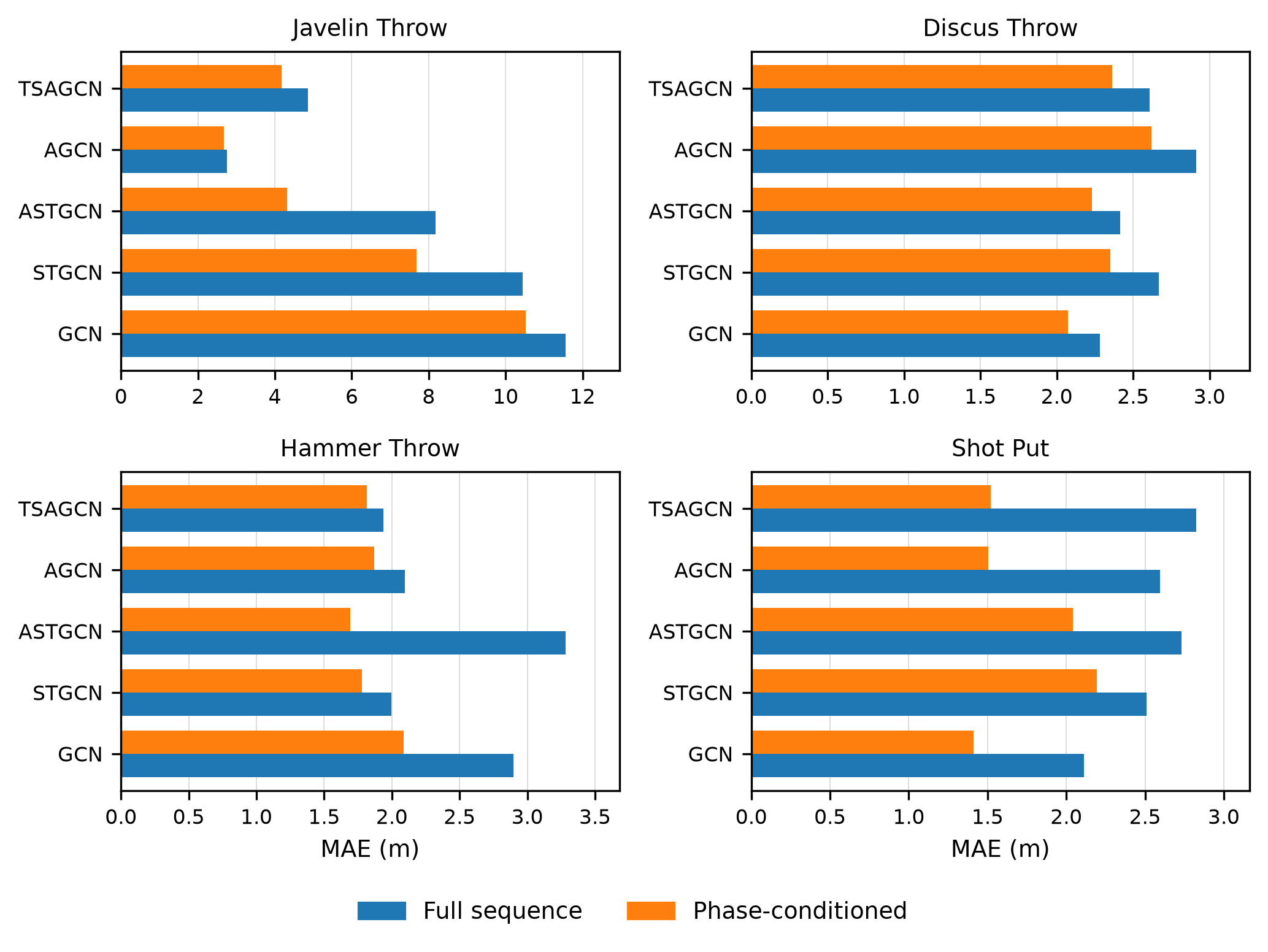}
\caption{MAE in meters for full-sequence and phase-conditioned score prediction across four throwing disciplines. Errors in meters are obtained by de-normalizing the normalized MAE using Eq.~\eqref{eq:mae_meter}.}
\label{fig:mae_meters_full_phase}
\end{figure}

Table~\ref{tab:mae_full_phase} reports the normalized mean absolute error (MAE) for full-sequence and phase-conditioned inputs across the four throwing disciplines, while Fig.~\ref{fig:mae_meters_full_phase} presents the corresponding errors after conversion to metres using Eq.~\eqref{eq:mae_meter}. Across all evaluated graph-based models and all four disciplines, the phase-conditioned setting produces lower prediction error than the corresponding full-sequence setting. For example, in javelin, the ASTGCN MAE decreases from $0.2023$ to $0.1070$, while in hammer it decreases from $0.2645$ to $0.1364$. The same overall pattern is observed in the metre-scale results shown in Fig.~\ref{fig:mae_meters_full_phase}.

The lowest phase-conditioned MAE differs across disciplines. AGCN obtains the lowest error for javelin ($0.0662$), GCN for discus ($0.1160$), ASTGCN for hammer ($0.1364$), and GCN for shot put ($0.2055$). Despite these differences in the best-performing backbone, the phase-conditioned input yields lower error than the corresponding full-sequence input for every graph-based model evaluated in Table~\ref{tab:mae_full_phase}. Additional repeated-seed and phase-boundary analyses are reported in Online Resource~1. The phase-boundary ablation compares the annotated four-phase partition (Manual-4)
with four equal temporal windows (Uniform-4) under the same AGCN backbone and five random seeds. Manual-4 gives the lower mean MAE for javelin ($\Delta = 0.019$ normalized, $0.783$~m) and is effectively tied with Uniform-4 for hammer ($\Delta = 0.001$ normalized, $0.014$~m), while Uniform-4 gives the lower mean MAE for discus ($\Delta = -0.008$) and shot put ($\Delta = -0.054$).

\sisetup{
  detect-weight=true,
  detect-inline-weight=math,
  table-number-alignment=center,
  round-mode=places,
  round-precision=4
}
\begin{table}[t]
\centering
\small
\caption{MAE in normalized space for full-sequence and phase-conditioned inputs across four throwing disciplines. Lower is better.}
\label{tab:mae_full_phase}
\setlength{\tabcolsep}{4pt}
\renewcommand{\arraystretch}{1.15}
\begin{tabular}{l
                S[table-format=1.4] S[table-format=1.4]
                S[table-format=1.4] S[table-format=1.4]
                S[table-format=1.4] S[table-format=1.4]
                S[table-format=1.4] S[table-format=1.4]}
\toprule
\multirow{2}{*}{\textbf{Model}} &
\multicolumn{2}{c}{\textbf{Javelin}} &
\multicolumn{2}{c}{\textbf{Discus}} &
\multicolumn{2}{c}{\textbf{Hammer}} &
\multicolumn{2}{c}{\textbf{Shot put}} \\
\cmidrule(lr){2-3}\cmidrule(lr){4-5}\cmidrule(lr){6-7}\cmidrule(lr){8-9}
& {\textbf{full}} & {\textbf{phase}}
& {\textbf{full}} & {\textbf{phase}}
& {\textbf{full}} & {\textbf{phase}}
& {\textbf{full}} & {\textbf{phase}} \\
\midrule
GCN \cite{kipf2016semi}        & 0.2863 & 0.2604 & 0.1278 & {\bfseries 0.1160} & 0.2334 & 0.1680 & 0.3076 & {\bfseries 0.2055} \\
STGCN \cite{yan2018stgcn}      & 0.2583 & 0.1902 & 0.1494 & 0.1317              & 0.1609 & 0.1432 & 0.3651 & 0.3196 \\
ASTGCN \cite{guo2019attention} & 0.2023 & 0.1070 & 0.1352 & 0.1249              & 0.2645 & {\bfseries 0.1364} & 0.3976 & 0.2972 \\
AGCN \cite{shi2019twostream}   & 0.0682 & {\bfseries 0.0662} & 0.1631 & 0.1467  & 0.1686 & 0.1504 & 0.3779 & 0.2191 \\
TSAGCN \cite{shi2020weakly}    & 0.1203 & 0.1034 & 0.1461 & 0.1324              & 0.1559 & 0.1461 & 0.4112 & 0.2213 \\
\bottomrule
\end{tabular}
\end{table}

\section{Discussion}
\label{sec:discussion}

Phase-conditioned modeling lowered throw-distance error in every discipline and under every
pose-graph backbone tested. The consistency matters more than the margin, because it
indicates that the gain comes from the representation rather than from a particular
architecture. A throw is a short sequence of functionally distinct actions, and a model
that pools an entire clip must average across them. Decomposing the sequence first lets
each segment contribute its own summary, so that information from the delivery is not
diluted by a long and variable approach. The metre-scale reduction is largest for javelin,
which fits this account, since the javelin run-up varies most across athletes and clips in
duration, rhythm, and scene composition~\cite{makino2023kinematic,makino2022kinematic}.

The phase-boundary ablation separates two things that the main result combines: segmenting
the sequence at all, and segmenting it at biomechanically aligned points. Both
formulations use four pooled descriptors, so any difference between them is attributable
to boundary placement alone. Little difference appears. For javelin, discus, and hammer the
gap between annotated and uniform boundaries falls within the across-seed variation, and
the one event where the gap exceeds seed variability, shot put, favours the uniform
windows. The reasonable conclusion is that temporal decomposition is the active ingredient
and that boundary placement is second-order at this sequence length and label volume. For
the rotational events this is expected, since discus and hammer turns blend into one
another and leave less for a boundary to align
to~\cite{leigh2010individualized,castaldi2022hammer}.

This has a practical reading for sports engineering. Competition broadcast footage is
easier to obtain than laboratory motion capture and closer to the conditions under which
elite sport is actually observed~\cite{wu2022survey,xu2025action}, and it carries enough
information for performance-oriented modeling once the representation is structured. The
phase annotations in Throwing4 earn their cost not by improving this particular regression
but by making a class of questions askable at all: which phase carries the performance
signal, whether boundaries can be recovered automatically, and how staged movement relates
to outcome across disciplines. Each of those requires labelled phases to study.

Four limitations bound these claims. The pose is two-dimensional and estimated from
broadcast video, so joint torques, implement dynamics, and true three-dimensional
kinematics remain out of reach. The data covers valid elite non-foul attempts, which
sharpens the labels but leaves generalization to novice athletes and training footage
untested. Phase boundaries are hand-annotated, so the phase-conditioned results are an
oracle-boundary benchmark rather than a deployable system. And because the clip-level
partition described in Section~\ref{subsec:evaluation} is not athlete-independent, these
numbers do not speak to athlete-independent generalization. The metre-scale error should
accordingly be read as coarse estimation, informative for attempt-level comparison and
technique analysis but not a substitute for official measurement.

Each limitation names its own next step: automatic phase detection to remove the oracle
boundaries, athlete-independent protocols to test transfer, phase-wise and joint-level
attribution to locate the signal within a throw, and multimodal or three-dimensional input
to recover the biomechanics that two-dimensional pose cannot.

\section{Conclusion}
\label{sec:conclusion}

This paper introduced Throwing4 as a resource for studying throw-distance prediction from elite athletics throwing under realistic broadcast conditions. The dataset combines curated RGB clips, identity-consistent MPII-16 pose sequences, frame-level phase annotations, and official throw distances across javelin, discus, hammer, and shot put. Using this resource, we formulated throw-distance prediction from pose as a regression problem and compared full-sequence modeling with a phase-conditioned alternative that explicitly respects the staged structure of throwing technique. Across all evaluated disciplines and graph-based backbones, phase-conditioned modeling produced lower prediction error than full-sequence regression under the present protocol. These results suggest that biomechanical phase structure is a useful inductive prior for pose-based performance modeling in elite athletics throwing and provide a basis for future work on phase-aware sports analytics from broadcast video.

\backmatter

\section*{Acknowledgements}
We acknowledge the Ministry of Youth Affairs and Sports, Department of Sports, Government of India, for
providing financial support to our research in Sports Analytics.

\section*{Declarations}

\subsection*{Conflict of Interest}
The authors declare that they have no conflict of interest.
\subsection*{Compliance with Ethical Standards}
This study did not involve the recruitment of human participants, nor any
intervention or direct experimentation. All video material was obtained from
publicly broadcast elite athletics competitions, and the work is limited to
secondary analysis of this existing media. Formal ethics approval and informed
consent were therefore not required for this study.



\section*{Data Availability}

The Throwing4 dataset and associated documentation are publicly available through the Throwing4 repository at
\url{https://github.com/Bikudebug/Throwing4-Dataset}.
The repository provides dataset documentation and access information for the RGB video clips, frame-level phase annotations, official throw-distance labels, and identity-consistent MPII-16 two-dimensional pose sequences used in this study.

\bibliography{sn-bibliography}

@inproceedings{niebles2010modeling,
  author    = {Niebles, Juan Carlos and Chen, Chao-Wei and Fei-Fei, Li},
  title     = {Modeling temporal structure of decomposable motion segments for activity classification},
  booktitle = {European Conference on Computer Vision (ECCV)},
  pages     = {392--405},
  publisher = {Springer},
  address   = {Berlin, Heidelberg},
  year      = {2010}
}

@inproceedings{jhuang2013towards,
  author    = {Jhuang, Hueihan and Gall, Juergen and Zuffi, Silvia and Schmid, Cordelia and Black, Michael J.},
  title     = {Towards understanding action recognition},
  booktitle = {Proceedings of the IEEE International Conference on Computer Vision (ICCV)},
  pages     = {3192--3199},
  year      = {2013}
}

@inproceedings{shao2020finegym,
  author    = {Shao, Dian and Zhao, Yue and Dai, Bo and Lin, Dahua},
  title     = {FineGym: A hierarchical video dataset for fine-grained action understanding},
  booktitle = {Proceedings of the IEEE/CVF Conference on Computer Vision and Pattern Recognition (CVPR)},
  pages     = {2616--2625},
  year      = {2020}
}

@article{wu2022survey,
  author  = {Wu, Fan and Wang, Qiang and Bian, Jia and Ding, Ning and Lu, Fei and Cheng, Jun and Dou, Dejing and Xiong, Hui},
  title   = {A survey on video action recognition in sports: Datasets, methods and applications},
  journal = {IEEE Transactions on Multimedia},
  volume  = {25},
  pages   = {7943--7966},
  year    = {2022}
}

@article{xu2025action,
  author  = {Xu, H. and Baniya, A. A. and Well, S. and Bouadjenek, M. R. and Dazeley, R. and Aryal, S.},
  title   = {Action spotting and precise event detection in sports: Datasets, methods, and challenges},
  journal = {arXiv preprint arXiv:2505.03991},
  year    = {2025}
}

@inproceedings{xu2022finediving,
  author    = {Xu, Jiazheng and Rao, Yujiu and Yu, Xulin and Chen, Guotai and Zhou, Jie and Lu, Jiwen},
  title     = {FineDiving: A fine-grained dataset for procedure-aware action quality assessment},
  booktitle = {Proceedings of the IEEE/CVF Conference on Computer Vision and Pattern Recognition (CVPR)},
  pages     = {2949--2958},
  year      = {2022}
}

@inproceedings{zhang2023logo,
  author    = {Zhang, S. and Dai, W. and Wang, S. and Shen, X. and Lu, J. and Zhou, J. and Tang, Y.},
  title     = {LOGO: A long-form video dataset for group action quality assessment},
  booktitle = {Proceedings of the IEEE/CVF Conference on Computer Vision and Pattern Recognition (CVPR)},
  pages     = {2405--2414},
  year      = {2023}
}

@inproceedings{li2021multisports,
  author    = {Li, Y. and Chen, L. and He, R. and Wang, Z. and Wu, G. and Wang, L.},
  title     = {MultiSports: A multi-person video dataset of spatio-temporally localized sports actions},
  booktitle = {Proceedings of the IEEE/CVF International Conference on Computer Vision (ICCV)},
  pages     = {13536--13545},
  year      = {2021}
}

@inproceedings{shi2020weakly,
  author    = {Shi, Baifeng and Dai, Qi and Mu, Yadong and Wang, Jingdong},
  title     = {Weakly-supervised action localization by generative attention modeling},
  booktitle = {Proceedings of the IEEE/CVF Conference on Computer Vision and Pattern Recognition (CVPR)},
  pages     = {1009--1019},
  year      = {2020}
}

@inproceedings{giancola2018soccernet,
  author    = {Giancola, Silvio and Amine, Mohieddine and Dghaily, Tarek and Ghanem, Bernard},
  title     = {SoccerNet: A scalable dataset for action spotting in soccer videos},
  booktitle = {Proceedings of the IEEE/CVF Conference on Computer Vision and Pattern Recognition (CVPR) Workshops},
  pages     = {1711--1721},
  year      = {2018}
}

@inproceedings{deliege2021soccernetv2,
  author    = {Deli{\`e}ge, Adrien and Cioppa, Anthony and Giancola, Silvio and Seikavandi, Mohammad Javad and Dueholm, Jonas V. and Nasrollahi, Kamal and Ghanem, Bernard and Moeslund, Thomas B. and Van Droogenbroeck, Marc},
  title     = {SoccerNet-v2: A dataset and benchmarks for holistic understanding of broadcast soccer videos},
  booktitle = {Proceedings of the IEEE/CVF Conference on Computer Vision and Pattern Recognition (CVPR) Workshops},
  year      = {2021}
}

@inproceedings{guo2019attention,
  author    = {Guo, Shengnan and Lin, Youfang and Feng, Ning and Song, Chao and Wan, Huaiyu},
  title     = {Attention based spatial-temporal graph convolutional networks for traffic flow forecasting},
  booktitle = {Proceedings of the AAAI Conference on Artificial Intelligence},
  volume    = {33},
  number    = {01},
  pages     = {922--929},
  year      = {2019}
}

@article{kipf2016semi,
  author  = {Kipf, Thomas N. and Welling, Max},
  title   = {Semi-supervised classification with graph convolutional networks},
  journal = {arXiv preprint arXiv:1609.02907},
  year    = {2016}
}

@inproceedings{badatya2025biomechanical,
  author    = {Badatya, Bikash Kumar and Baghel, V. and Amin, Jyotirmoy and Hegde, Ravi},
  title     = {Biomechanical-phase based temporal segmentation in sports videos: A demonstration on Javelin-throw},
  booktitle = {2025 IEEE International Workshop on Sport, Technology and Research (STAR)},
  pages     = {19--24},
  year      = {2025}
}

@inproceedings{yan2018stgcn,
  author    = {Yan, Sijie and Xiong, Yuanjun and Lin, Dahua},
  title     = {Spatial Temporal Graph Convolutional Networks for Skeleton-Based Action Recognition},
  booktitle = {Proceedings of the AAAI Conference on Artificial Intelligence},
  year      = {2018}
}

@inproceedings{shi2019twostream,
  author    = {Shi, Lei and Zhang, Yifan and Cheng, Jian and Lu, Hanqing},
  title     = {Two-stream adaptive graph convolutional networks for skeleton-based action recognition},
  booktitle = {Proceedings of the IEEE/CVF Conference on Computer Vision and Pattern Recognition (CVPR)},
  pages     = {12026--12035},
  year      = {2019}
}

@inproceedings{liu2020msg3d,
  author    = {Liu, Ziyu and Zhang, Hongwen and Chen, Zhenghao and Wang, Zhiyong and Ouyang, Wanli},
  title     = {Disentangling and unifying graph convolutions for skeleton-based action recognition},
  booktitle = {Proceedings of the IEEE/CVF Conference on Computer Vision and Pattern Recognition (CVPR)},
  pages     = {143--152},
  year      = {2020}
}

@inproceedings{chen2021ctrgcn,
  author    = {Chen, Yuxin and Zhang, Ziqi and Yuan, Chunfeng and Li, Bing and Deng, Ying and Hu, Weiming},
  title     = {Channel-wise topology refinement graph convolution for skeleton-based action recognition},
  booktitle = {Proceedings of the IEEE/CVF International Conference on Computer Vision (ICCV)},
  year      = {2021}
}

@inproceedings{cao2017openpose,
  author    = {Cao, Zhe and Simon, Tomas and Wei, Shih-En and Sheikh, Yaser},
  title     = {Realtime multi-person 2D pose estimation using part affinity fields},
  booktitle = {Proceedings of the IEEE Conference on Computer Vision and Pattern Recognition (CVPR)},
  year      = {2017}
}

@article{soomro2012ucf101,
  author  = {Soomro, Khurram and Zamir, Amir Roshan and Shah, Mubarak},
  title   = {UCF101: A dataset of 101 human action classes from videos in the wild},
  journal = {arXiv preprint arXiv:1212.0402},
  year    = {2012}
}

@article{kay2017kinetics,
  author  = {Kay, Will and Carreira, Joao and Simonyan, Karen and Zhang, Brian and Hillier, Chloe and Vijayanarasimhan, Sudheendra and Viola, Fabio and Green, Tim and Back, Trevor and Natsev, Paul and Zisserman, Andrew},
  title   = {The Kinetics human action video dataset},
  journal = {arXiv preprint arXiv:1705.06950},
  year    = {2017}
}

@misc{mmpose2020,
  author       = {{MMPose Contributors}},
  title        = {MMPose: OpenMMLab pose estimation toolbox and benchmark},
  year         = {2020},
  note         = {arXiv:2011.00357},
  howpublished = {\url{https://github.com/open-mmlab/mmpose}}
}

@inproceedings{kukleva2019unsupervised,
  author    = {Kukleva, Anna and Kuehne, Hilde and Sener, Fadime and Gall, Juergen},
  title     = {Unsupervised learning of action classes with continuous temporal embedding},
  booktitle = {Proceedings of the IEEE/CVF Conference on Computer Vision and Pattern Recognition (CVPR)},
  pages     = {12066--12074},
  year      = {2019}
}

@inproceedings{kumar2022unsupervised,
  author    = {Kumar, S. and Haresh, S. and Ahmed, A. and Konin, A. and Zia, M. Z. and Tran, Q.-H.},
  title     = {Unsupervised action segmentation by joint representation learning and online clustering},
  booktitle = {Proceedings of the IEEE/CVF Conference on Computer Vision and Pattern Recognition (CVPR)},
  pages     = {20174--20185},
  year      = {2022}
}

@inproceedings{xu2024temporally,
  author    = {Xu, M. and Gould, S.},
  title     = {Temporally consistent unbalanced optimal transport for unsupervised action segmentation},
  booktitle = {Proceedings of the IEEE/CVF Conference on Computer Vision and Pattern Recognition (CVPR)},
  pages     = {14618--14627},
  year      = {2024}
}

@article{zaras2021biological,
  author  = {Zaras, Nikolaos and Stasinaki, Athanasia-Nikoletta and Methenitis, Spyridon and Krase, Anastasia and Karampatsos, Georgios and Georgiadis, Georgios and Spengos, Konstantinos and Terzis, Gerasimos},
  title   = {Biological Determinants of Track and Field Throwing Performance},
  journal = {Journal of Functional Morphology and Kinesiology},
  volume  = {6},
  number  = {2},
  pages   = {40},
  year    = {2021}
}

@article{makino2023kinematic,
  author  = {Makino, Mizuki and Nakayama, Koichi and Ando, Yuka and Tauchi, Kenji},
  title   = {Kinematic Contribution to Javelin Velocity at Different Run-Up Velocities in Male Athletes},
  journal = {Journal of Human Kinetics},
  volume  = {89},
  pages   = {43--52},
  year    = {2023}
}

@article{makino2022kinematic,
  author  = {Makino, Mizuki and Tauchi, Kenji},
  title   = {Kinematic Factors Related to Forward and Vertical Release Velocity in Male Javelin Throwers},
  journal = {International Journal of Sport and Health Science},
  volume  = {20},
  pages   = {249--259},
  year    = {2022}
}

@article{leigh2010individualized,
  author  = {Leigh, Steve and Liu, Hui and Hubbard, Mont and Yu, Bing},
  title   = {Individualized Optimal Release Angles in Discus Throwing},
  journal = {Journal of Biomechanics},
  volume  = {43},
  number  = {3},
  pages   = {540--545},
  year    = {2010}
}

@article{castaldi2022hammer,
  author  = {Castaldi, Giulio M. and Borzuola, Riccardo and Giombini, Arrigo and Fischetti, Francesco and Macaluso, Andrea},
  title   = {Biomechanics of the Hammer Throw: Narrative Review},
  journal = {Frontiers in Sports and Active Living},
  volume  = {4},
  pages   = {853536},
  year    = {2022}
}

@article{mastalerz2022shotput,
  author  = {Mastalerz, Andrzej and Sadowski, Jerzy},
  title   = {Variability of Performance and Kinematics of Different Shot Put Techniques in Elite and Sub-Elite Athletes---A Preliminary Study},
  journal = {International Journal of Environmental Research and Public Health},
  volume  = {19},
  number  = {3},
  pages   = {1751},
  year    = {2022}
}

\end{document}